\documentclass{article}
\usepackage{spconf,amsmath,graphicx}
\usepackage{amsfonts}
\usepackage{booktabs}

\usepackage{enumitem}
\setlist{nosep, leftmargin=14pt}

\usepackage{mwe} % to get dummy images
\usepackage{array} % 记得加这个

\newcolumntype{C}[1]{>{\centering\arraybackslash}m{#1}}

\title{See in Depth: Training-Free Surgical Scene Segmentation \\with Monocular Depth Priors}
\name{Kunyi Yang$^{*}$, Qingyu Wang$^{*}$, Cheng Yuan, Yutong Ban \thanks{$*$ denotes equal contributions}}
\address{Global College, Shanghai Jiao Tong University, Shanghai, China}
\begin{document}
%\ninept
%
\maketitle
\begin{abstract}
Pixel-wise segmentation of laparoscopic scenes is essential for computer-assisted surgery but difficult to scale due to the high cost of dense annotations. We propose depth-guided surgical scene segmentation (DepSeg), a training-free framework that utilizes monocular depth as a geometric prior together with pretrained vision foundation models. DepSeg first estimates a relative depth map with a pretrained monocular depth estimation network and proposes depth-guided point prompts, which SAM2 converts into class-agnostic masks. Each mask is then described by a pooled pretrained visual feature and classified via template matching against a template bank built from annotated frames. On the CholecSeg8k dataset, DepSeg improves over a direct SAM2 auto segmentation baseline (35.9\% vs.\ 14.7\% mIoU) and maintains competitive performance even when using only 10--20\% of the object templates. These results show that depth-guided prompting and template-based classification offer an annotation-efficient segmentation approach.

\end{abstract}
\begin{keywords}
Scene Segmentation, Surgical Scene Segmentation, Medical Segmentation\end{keywords}

\section{Introduction}

Accurate pixel-wise segmentation of {surgical} scenes in laparoscopic videos is crucial for instrument tracking, workflow analysis, and context-aware assistance in the operating room. However, obtaining dense ground-truth masks is expensive and time-consuming, as it requires expert annotators and careful delineation of instruments and tissue across thousands of frames. In practice, hospitals typically have access to abundant video but only a {small number of annotations}, which makes fully supervised training and maintenance of large segmentation networks difficult in real deployments.

Recently, vision {foundation models} are developing rapidly. Two representative families are the SAM ~\cite{kirillov2023segment} series of promptable segmentation models and DINO series self-supervised encoders~\cite{caron2021emerging}. SAM2~\cite{ravi2024sam} provides strong, class-agnostic masks from simple prompts. It has already been adapted to surgical data: Surgical SAM2~\cite{liu2024surgical} proposes a real-time SAM2-based pipeline for surgical video segmentation by optimizing memory and computation for high frame rates. At the same time, MedSAM2~\cite{ma2025medsam2} utilizes SAM2 as a backbone for general medical image and video segmentation, demonstrating robust performance. Other studies systematically benchmark SAM2 across diverse surgical datasets and prompting strategies, highlighting its zero-shot robustness but also observing that it remains inherently class-agnostic~\cite{yuan2024segment}. DINO encoders~\cite{oquab2023dinov2,simeoni2025dinov3} produce rich visual features that transfer well across domains. For example, CNOS~\cite{nguyen2023cnos} combines SAM mask proposals with DINOv2 descriptors and CAD templates to perform training-free novel object segmentation via feature matching. These works suggest a natural division of roles: SAM2 for generating high-quality, class-agnostic masks, and DINO-style encoders for assigning semantic labels via template banks.

Geometric cues, particularly depth information, play an important role in understanding surgical scenes. In laparoscopic imagery, instruments are generally positioned closer to the camera than the surrounding tissue, while different anatomical layers appear at distinct depth levels, collectively forming a structured spatial layout. Motivated by this observation, we incorporate depth as a lightweight geometric prior to guide segmentation, encouraging the generation of coherent, object-like regions. Pre-trained depth estimation models, such as DepthAnything~\cite{yang2024depth}, cannot yield precise metric geometry. But it reliably captures relative spatial relationships within the scene, which can be useful in further scene understanding tasks.

Building on these ideas, we propose {DepSeg}, a training-free framework for surgical instrument and tissue segmentation which combines depth cues with frozen foundation models SAM2 and DINOv3. From monocular {depth} estimates, we propose the point prompts with guidance from depth, which SAM2 converts into class-agnostic masks. Each mask is then described by a pooled DINOv3 token and classified via template matching with a pre-built template bank. Our main contributions are:
\begin{itemize}
    \item We propose an annotation-efficient segmentation framework that leverages pretrained vision {foundation models}, relying only on a template bank constructed from a {low number of annotations}, without any fine-tuning or additional learnable components.
    
    \item We propose a depth-guided prompting strategy for laparoscopic {surgical} scenes, where monocular {depth} is  used as a prior to aid segmentation by suggesting coherent, object-like regions.
    \item We highlight two practical advantages of this design: it is {training-free}, requiring only forward passes and feature template storage, and it is {easy to expand}, since new annotated frames or classes can be incorporated simply by adding templates to the bank.
    % \item We demonstrate experimentally that combining depth priors, DepthAnythingV2, SAM2-based mask proposals, and DINOv3-based template matching yields competitive instrument and tissue segmentation performance on laparoscopic surgical benchmarks, while keeping annotation and maintenance costs low.
\end{itemize}

\section{Method}
\begin{figure}
    \centering
\includegraphics[width=1\linewidth]{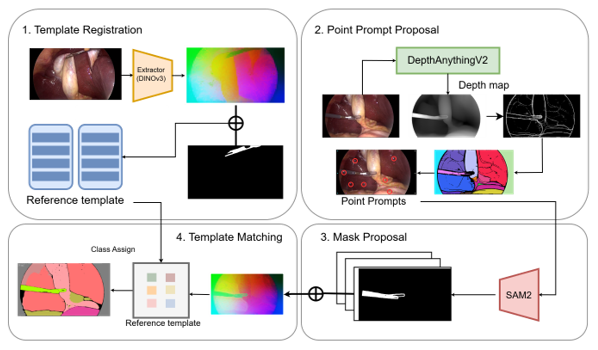}
    \caption{The proposed DepSeg Pipeline}
    \label{fig:placeholder}
\end{figure}

The proposed depth-guided segmentation (DepSeg) pipeline consists of four parts. First, the Point Prompt Proposal generates geometric prompts from an RGB frame. Then, the Mask Proposal Module uses these prompts with SAM2 to create refined, class-agnostic masks. These masks are then classified with the Template Matching by matching DINOv3 descriptors against a template bank, which is built offline during the Template Registration Stage~(\ref{sec:template_registration}). A final small-area-first merge produces the pixel-wise segmentation map.

\subsection{Problem formulation}

Let $\Omega = \{1,\dots,H\} \times \{1,\dots,W\}$ denote the image domain and
$x \in \mathbb{R}^{H \times W \times 3}$ a single RGB laparoscopic frame.
We consider a finite set of foreground classes $\mathcal{C}$ that includes
instrument and tissue categories, and define the pixelwise label space
$\mathcal{Y} = \{0\} \cup \mathcal{C}$, where $0$ denotes background. The objective is to predict an instrument/tissue/background label for every pixel:
\[
F : \mathbb{R}^{H \times W \times 3} \to \mathcal{Y}^{H \times W},
\]
where $y = F(x)$ is the final segmentation map.

\subsection{Template Registration}
\label{sec:template_registration}

This offline stage aims to generate a template bank of class-representative descriptors. The process uses a labeled training set, denoted as $\{(x^{(n)}, y^{(n)})\}_{n=1}^N$, where $N$ is the total number of training samples, $x^{(n)}$ is the $n$-th training RGB image, and $y^{(n)}$ is its corresponding pixel-wise ground-truth label map. For each training image $x^{(n)}$, a frozen DINOv3 encoder, produces a token grid $T \in \mathbb{R}^{H' \times W' \times D}$. This grid is a collection of $D$-dimensional feature vectors arranged spatially with dimensions $H' \times W'$.
For each semantic class $c$ from the set of all possible labels $\mathcal{Y}$, a binary mask $M_{n,c}$ is created from the ground-truth map $y^{(n)}$. This mask identifies all pixels belonging to class $c$:
\[
 M_{n,c}(p) = \mathbf{1}[y^{(n)}(p) = c], \quad p \in \Omega,
\]
where  $\mathbf{1}[\cdot]$ is the indicator function, which is 1 if the condition inside is true and 0 otherwise. Such a mask is downsampled to the token grid's resolution ($H' \times W'$) to create a new mask $M'_{n,c}$. A mask-aware pooled descriptor $h_{n,c}$ is then computed by averaging the token vectors over the spatial locations $(u,v)$ covered by the downsampled mask:
\[
 h_{n,c} = \frac{1}{Z_{n,c}} \sum_{u,v} M'_{n,c}(u,v)\, T(u,v,:) \in \mathbb{R}^D,
\]
where $T(u,v,:)$ refers to the $D$-dimensional feature vector at grid location $(u,v)$, and the normalization factor $Z_{n,c} = \sum_{u,v} M'_{n,c}(u,v)$ is the area of the downsampled mask. The resulting descriptor $h_{n,c}$ is then L2-normalized. Finally, all descriptors for a given class $c$ are collected from the entire training set to form a template bank, $\mathcal{T}_c = \{ t_{c,\ell} \}_{\ell=1}^{L_c}$. Here,  each $t_{c,\ell}$ is a normalized descriptor vector. This bank is saved and loaded during the inference phase.

\subsection{Depth-Guided Point Prompt Proposal}
\label{sec:point_proposal}

This stage takes a single input RGB frame $x \in \mathbb{R}^{H \times W \times 3}$ and generates a sparse set of high-quality point prompts. First, a pretrained monocular depth estimator, DepthAnythingV2\cite{yang2024depth}, produces a dense depth field $d \in \mathbb{R}^{H \times W}$. From this, a relative depth map $d_{\text{rel}}$ is calculated by subtracting the median depth value of the entire image.
Using $d_{\text{rel}}$ and $x$, we apply a deterministic proposal operator $G$ composed of K-means clustering~\cite{macqueen1967multivariate}, Canny edge detection~\cite{canny2009computational}, Otsu thresholding~\cite{otsu1975threshold}, and a watershed transform~\cite{vincent1991watersheds} on the depth map to generate a set of initial region proposals $\mathcal{R} = \{ R_m \}_{m=1}^M$, where $R_m \subseteq \Omega$ is a single region proposal and $M$ is the total number of proposals.
 These proposals are regularized using morphological operators to yield cleaned regions $\tilde{R}_m$.
For each cleaned region $\tilde{R}_m$, a distance transform is computed:
\[
D_m(p) = \operatorname{dist}(p, \partial \tilde{R}_m), \qquad p \in \Omega,
\]
where $\operatorname{dist}$ is a distance metric (e.g., Euclidean), and $\partial \tilde{R}_m$ is the boundary of region $\tilde{R}_m$. A small set of interior point prompts, denoted $S_m$, is obtained by detecting local maxima of the distance map $D_m(p)$, subject to constraints on minimum distance and border exclusion.

\subsection{Mask Proposal}
\label{sec:mask_proposal}

This stage uses the point prompts $\{S_m\}$ to produce a set of refined, class-agnostic object masks. A frozen SAM2\cite{ravi2024sam} image predictor takes the RGB image $x$ and the prompts $S_m$ for each proposed region to generate a mask score map. These score maps are thresholded to obtain initial binary masks, and then cleaned with a morphological opening operator. Masks with an area smaller than a minimum threshold are discarded.
To resolve overlaps, the kept masks are merged into a priority label map $L$. The masks are sorted by ascending area, and each pixel is assigned the ID of the first (i.e., smallest) mask that covers it. This process preserves the details of smaller objects like instruments.
Next, the merged map is decomposed to ensure each resulting mask is a single, contiguous region. For each original object ID in $L$, we find all its spatially connected components. Each component is then treated as a separate binary mask, which will forms then final set of refined masks $\widehat{\mathcal{M}} = \{ M_j \mid j = 1,\dots,J \}$, where $J$ is the number of final masks.

\subsection{Template Matching}
\label{sec:template_matching}

In this final stage, each refined mask $M_j \in \widehat{\mathcal{M}}$ is classified and merged to produce the final segmentation map. First, a DINOv3 descriptor $h_j$ is computed for each mask using the same mask-aware pooling process from the Template Registration Stage. The mask $M_j$ is downsampled to the token grid resolution to create $M'_j$, and the descriptor is computed and L2-normalized:
\[
 h_j = \operatorname{norm}\left( \frac{1}{Z_j} \sum_{u,v} M'_j(u,v) \, T(u,v,:) \right) \in \mathbb{R}^D,
\]
where $Z_j$ is the area of the downsampled mask $M'_j$.
Next, each descriptor $h_j$ is compared against the pre-computed template banks $\{\mathcal{T}_c\}$. The cosine similarity between the descriptor and each template vector is calculated:
\[
s_{j,c,\ell} = h_j^\top t_{c,\ell},
\]
For each class $c$, a score $S_{j,c}$ is computed by summing the top-$k$ similarity values, where $k$ is a hyperparameter:
\[
S_{j,c} = \sum_{\ell \in \operatorname{TopK}_\ell(s_{j,c,\ell})} s_{j,c,\ell}.
\]
Here, $\operatorname{TopK}_\ell(s_{j,c,\ell})$ denotes the set of indices $\ell$ corresponding to the $k$ largest similarity values for class $c$. The mask $M_j$ is assigned the class $\hat{c}_j$ with the highest score:
\[
\hat{c}_j = \arg\max_{c \in \mathcal{Y}} S_{j,c}.
\]
Finally, the set of labeled masks $\{ (M_j, \hat{c}_j) \}$ is sorted by ascending area. A final label map $y$ is generated by iterating through the sorted masks, where each mask assigns its class label only to previously unlabeled pixels. This small-area-first merge strategy yields the final pixel-wise prediction $y = F(x)$ and keeps the details of small objects.

\section{Experiments}

\textbf{Dataset} CholecSeg8k dataset~\cite{CholecSeg8k} contains 8{,}080 laparoscopic
cholecystectomy image frames from 17 video clips in Cholec80. Each frame
in the dataset is covered by three masks: one color mask, one mask used
by the annotation tool and one watershed mask. We split the dataset into
a training set and an evaluation set at approximately a 3:1 ratio.

\textbf{Settings} All experiments are conducted on a workstation with an NVIDIA RTX~A6000
GPU (48\,GB VRAM) and dual AMD EPYC~7763 64-core CPUs, running Ubuntu
22.04.4~LTS. Python~3.13.7 and PyTorch~2.8.0 are used for all
implementations. DepthAnythingV2, SAM2, and DINOv3 are employed as
frozen backbones

 \begin{table*}[t]
    \centering
    \resizebox{\textwidth}{!}{
    \begin{tabular}{l*{14}{C{1.5cm}}}
  \toprule
  & \multicolumn{14}{c}{CholecSeg8k IoU (\%)} \\
  \cmidrule(lr){2-15}
  & Background & Abdominal Wall & Liver & Gastrointestinal Tract & Fat & Grasper &
  Connective Tissue & Blood & Cystic Duct & L-hook Electrocautery & Gallbladder & Hepatic Vein & Liver
  Ligament & mIoU \\
  \midrule
    \multicolumn{15}{l}{\textbf{Fully Supervised Methods}} \\
    Mask2Former~\cite{cheng2022masked} & 98.0 & 92.1 & 92.5 & 80.7 & 90.0 & 84.8 &78.8 & 61.3 & 65.7 &91.3 & 88.6 & 0.0 & 95.8 & 78.4 \\
    \midrule
    \multicolumn{15}{l}{\textbf{Baseline}} \\
   SAM2 (auto) + Cosine Matching & 46.6 & 14.3 & 14.8 & 10.5 & \underline{60.7} & 2.4 & 1.1 & 0.1 & 0.0 & 5.9 & 35.2 & 0.0 & 0.0 & 14.7 \\
  \midrule
  \multicolumn{15}{l}{\textbf{Ablation on Different Modules}} \\
  SAM2 (auto) + Template Matching (ours) & 87.0 & 41.7 & 42.7 & 17.3 & 38.8 & \textbf{74.9} & 3.8 & \textbf{6.8} &
  0.0 & \underline{55.4} & 42.0 & 0.0 & 0.0 & 31.6 \\
  Depth Prompt (ours) + Cosine Matching & 55.7 & 5.2 & 13.6 & 12.5 & 56.5 & 5.3 & 4.9 & 0.1 & 0.0 & 12.0 & 40.7 & 0.0 & 0.0 & 15.9 \\
  \midrule
  \multicolumn{15}{l}{\textbf{Ablation on Template Matching Fractions}} \\
  Frac = 0.01 & \underline{94.8} & 61.3 & 61.8 & 0.0 & \textbf{61.9} & 59.3 & 0.0 & 0.0 & 0.0 & 0.0 & 42.8 & 0.0 & 0.0 & 29.4 \\
  Frac = 0.1  & \underline{94.8} & 61.0 & 66.7 & \underline{29.3} & {54.6} & 56.9 & 0.0 & 0.0 & 0.0 & 0.0 & 42.3 & 0.0 & 0.0 & 31.2 \\
  Frac = 0.2  & \textbf{94.9} & \underline{67.6} & 70.3 & \textbf{31.0} & 47.3 & 57.7 & 6.0 & 3.5 & 0.0 & 13.1 & \underline{45.6} & 0.0 & 0.0 & 33.6 \\
  Frac = 0.5  & \textbf{94.9} & 67.5 & 70.2 & 23.8 & 41.3 & 61.0 & 5.9 & \underline{6.0} & 0.0 & 29.7 & 44.7 & 0.0 & 0.0 & \underline{34.2} \\
  Frac = 0.8  & \underline{94.8} & 67.0 & 70.2 & 22.1 & 38.4 & 61.1 & 6.0 & 5.7 & 0.0 & 31.0 & 44.4 & 0.0 & 0.0 & 33.9 \\
    Frac = 1.0  & 93.2 & 66.9 & 70.9 & 22.6 & 38.8 & 64.4 & \underline{6.1} & 4.4 & 0.0 & 54.9 & 44.9 & 0.0 & 0.0 & \textbf{35.9} \\
  \midrule
  \multicolumn{15}{l}{\textbf{Ablation on Top-$k$ similarities ($k$)}} \\
  Top-$k$ (k = 3)  & 46.2 & 43.5 & \underline{71.0} & 21.2 & 37.8 & \underline{66.1} & \textbf{6.2} & 4.2 & 0.0 & \textbf{69.2} & 44.5 & 0.0 & 0.0 & 31.5 \\
  Top-$k$ (k = 7)  & 93.2 & 66.9 & 70.9 & 22.6 & 38.8 & 64.4 & \underline{6.1} & 4.4 & 0.0 & 54.9 & 44.9 & 0.0 & 0.0 & \textbf{35.9} \\
  Top-$k$ (k = 15) & 94.7 & \textbf{68.8} & \textbf{71.1} & 24.0 & 41.0 & 60.9 & 6.0 & 3.2 & 0.0 & 27.2 & \textbf{45.7} & 0.0 & 0.0 & 34.0 \\
  \bottomrule
\end{tabular}
    }
    \caption{Per-class IoU and mean IoU (mIoU) on CholecSeg8k for automatic SAM2 mask generation
  (auto\_sam), different fractions of templates used (Frac), and different numbers of top-$k$ similarity values. Frac denotes the fraction of templates used in the process; in the Top-$k$ rows, $k$ is the number of highest similarity scores aggregated per location.Automatic means using SAM2 to generate masks directly instead of depth-based point prompts.}
    \label{tab:cholecseg8k_iou}
  \end{table*}

\textbf{Results} Table~\ref{tab:cholecseg8k_iou} reports per-class IoU and mean IoU (mIoU) on CholecSeg8k. The fully supervised baseline (Mask2Former) achieves an mIoU of 78.4\% when trained end-to-end on dense pixel labels. Our method is completely training-free. All the components SAM2, DINOv3, and DepthAnythingV2 remain frozen and only a template bank is constructed from the labeled training images.

We first consider a baseline, namely cosine Matching, which uses region position and area as features. For each mask, we compute an 8-D descriptor consisting of mean RGB (3), RGB standard deviation (3), area ratio (1), and aspect ratio (1). During training, we aggregate these descriptors per class into class templates. At test time, we compute the same 8-D descriptor for each mask and assign the class of the nearest prototype using cosine similarity. Applying this to masks from SAM2 ({SAM2 auto + cosine Matching}) gives 14.7\% mIoU providing a lower bound .

Comparing the{SAM2 auto + DINOv3 Template Matching} baseline to DepSeg with full templates (Frac = 1.0), we observe an improvement in mIoU from 31.6\% to 35.9\%. DepSeg also produces higher IoUs on several key tissue classes, such as{Abdominal Wall} (41.7\% to 66.9\%) and{Liver} (42.7\% to 70.9\%). For the instrument class{L-hook Electrocautery}, using $k=3$ in the top-$k$ aggregation reaches 69.2\% IoU, compared to 55.4\% for the automatic SAM2 baseline.

The effect of the template fraction on performance shows a clear trend. When only 1\% of the templates are used (Frac = 0.01), the mIoU reaches 29.4\%, and performance steadily increases up to 35.9\% at Frac = 1.0. Several low fractions remain competitive: using 10--20\% of the templates (Frac = 0.1/0.2) already yields 31.2--33.6\% mIoU, and with 50\% of the templates (Frac = 0.5) the mIoU (34.2\%) is within about 1.5\% of the full-template setting. The top-$k$ ablation indicates that a moderate value ($k=7$) provides the best overall mIoU (35.9\%), while smaller ($k=3$) or larger ($k=15$) values slightly degrade the balance between foreground and background performance.

\begin{figure}
    \centering
    \includegraphics[width=0.9\linewidth]{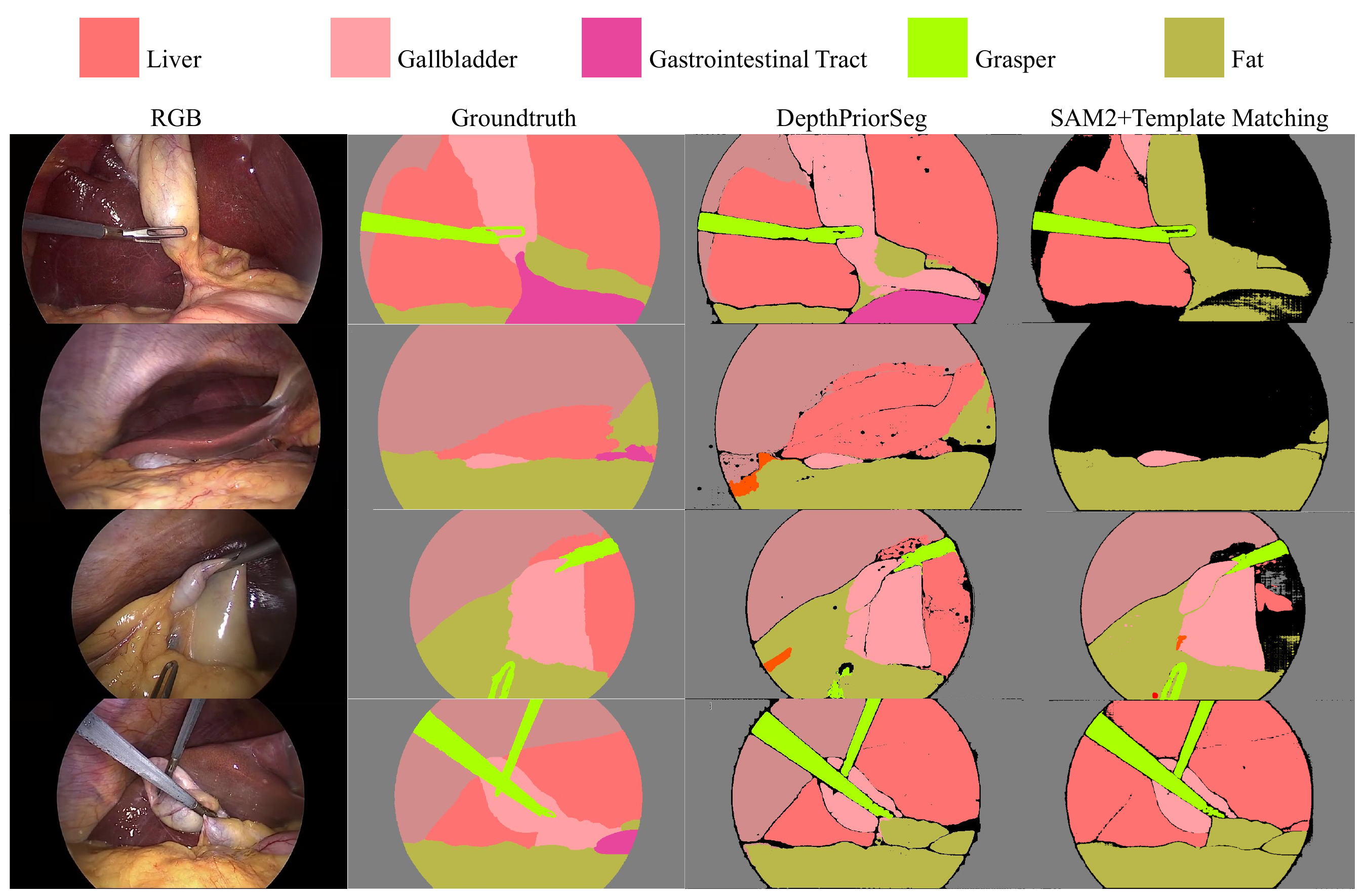}
    \caption{Qualitative results on CholecSeg8k. From left to right: input image, ground truth, DepSeg predictions, and SAM2 + Template Matching predictions.}
    \label{fig:vis}
\end{figure}

Qualitative examples in Figure~\ref{fig:vis} demonstrate the results from both DepSeg and the baseline SAM2 + DINOv3 Template Matching. Although the SAM2 automatic mask generator can produce masks with smoother boundaries between different objects, it is hard for the generator to propose all the object-like regions, leading to massive unsegmented areas. In contrast, DepSeg recovers more complete organ regions, especially around the gallbladder and liver.

Overall, the results indicate that depth-guided prompting offers a clear advantage over SAM2 automatic mask generation. Using depth maps from DepthAnythingV2 to propose point prompts helps SAM2 focus on object regions, which leads to more complete segmentation of instruments and tissues. The template fraction experiments show that our framework is annotation-efficient: performance improves steadily with more templates, and a relatively small subset (10--20\%) can achieve most of the final mIoU. The top-$k$ study shows a trade-off in similarity aggregation: very small $k$ values emphasize a few templates and reduce robustness, while very large $k$ values dilute the discriminative signal; a moderate choice ($k=7$ in our experiments) offers a good balance. The remaining gap to the fully supervised baseline is mainly attributable to small or rare classes, suggesting future work on better template selection, class balancing, and potentially incorporating temporal information.

\section{Conclusion}

We presented DepSeg, a training-free framework for surgical segmentation that combines monocular depth from DepthAnythingV2, depth-guided prompts for SAM2, and DINOv3-based template matching. On CholecSeg8k, the method improves over a direct SAM2 baseline and scales smoothly with the number of templates, while requiring only forward passes on frozen foundation models. This makes DepSeg an annotation-efficient and easily extendable alternative to fully supervised networks, with future extensions targeting temporal consistency and handling of rare classes.

\bibliographystyle{IEEEbib}
\bibliography{strings,refs}

\end{document}